\documentclass[a4paper,fleqn]{cas-dc}

\usepackage[round]{natbib}
\usepackage{subcaption}
\usepackage{pifont}
\usepackage{bm}
\newcommand{\xmark}{\ding{55}}%
\usepackage{tikz}
\usetikzlibrary{spy}
\usepackage{tabu}

\def\tsc#1{\csdef{#1}{\textsc{\lowercase{#1}}\xspace}}
\tsc{WGM}
\tsc{QE}

\usepackage{color}
\usepackage{cancel}

\newcommand{\ie}[1]{{i.e.}}
\newcommand{\etal}[1]{{et al.}}

\begin{document}
\let\WriteBookmarks\relax
\def\floatpagepagefraction{1}
\def\textpagefraction{.001}
\captionsetup[figure]{labelfont={bf},labelformat={default},labelsep=period,name={Fig.}}

\shorttitle{GLF-CR}    

\shortauthors{Fang Xu, et al}  

\title [mode = title]{GLF-CR: SAR-Enhanced Cloud Removal with Global-Local Fusion}  



%

\author[1,2]{Fang Xu}[type=editor,
			          auid=1,
			          bioid=1,
			          prefix=,]
\ead{xufang@whu.edu.cn}

\author[3]{Yilei Shi}[type=editor,
					  auid=2,
					  bioid=2,
					  prefix=,]
\ead{yilei.shi@tum.de}

\author[2]{Patrick Ebel}[type=editor,
					  auid=3,
					  bioid=3,
					  prefix=,]
\ead{patrick.ebel@tum.de}
					  
\author[1]{Lei Yu}[type=editor,
					 auid=4,
					 bioid=4,
					 prefix=,]
\ead{ly.wd@whu.edu.cn}

\author[4]{Gui-Song Xia}[type=editor,
					 auid=6,
					 bioid=6,
					 prefix=,]
\ead{guisong.xia@whu.edu.cn}
					 
\author[1]{Wen Yang}[type=editor,
					 auid=5,
					 bioid=5,
					 prefix=,]
\ead{yangwen@whu.edu.cn}
\cormark[1]
					 
\author[2,5]{Xiao Xiang Zhu}[type=editor,
					 auid=7,
					 bioid=7,
					 prefix=,]
\ead{xiaoxiang.zhu@dlr.de}
\cormark[1]






\affiliation[1]{organization={School of Electronic Information, Wuhan University},
	            city={Wuhan},
	            postcode={430072},
	            country={China}}
        
\affiliation[2]{organization={Data Science in Earth Observation, Technical University of Munich},
				city={Munich},
				postcode={80333},
				country={Germany}}
				
\affiliation[3]{organization={Remote Sensing Technology, Technical University of Munich},
				city={Munich},
				postcode={80333},
				country={Germany}}
				
\affiliation[4]{organization={School of  Computer Science, Wuhan University},
	            city={Wuhan},
	            postcode={430072},
	            country={China}}
				
\affiliation[5]{organization={Remote Sensing Technology Institute, German Aerospace Center},
				city={Weßling},
				postcode={82234},
				country={Germany}}






\cortext[1]{Corresponding author}



\begin{abstract}
	The challenge of the cloud removal task can be alleviated with the aid of Synthetic Aperture Radar (SAR) images that can penetrate cloud cover. However, the large domain gap between optical and SAR images as well as the severe speckle noise of SAR images may cause significant interference in SAR-based cloud removal, resulting in performance degeneration. In this paper, we propose a novel global-local fusion based cloud removal (GLF-CR) algorithm to leverage the complementary information embedded in SAR images. Exploiting the power of SAR information to promote cloud removal entails two aspects. The first, global fusion, guides the relationship among all local optical windows to maintain the structure of the recovered region consistent with the remaining cloud-free regions. The second, local fusion, transfers complementary information embedded in the SAR image that corresponds to cloudy areas to generate reliable texture details of the missing regions, and uses dynamic filtering to alleviate the performance degradation caused by speckle noise.
	Extensive evaluation demonstrates that the proposed algorithm can yield high quality cloud-free images and outperform state-of-the-art cloud removal algorithms with a gain about $1.7$ dB in terms of PSNR on SEN12MS-CR dataset.
\end{abstract}



\begin{keywords}
 Cloud removal \sep Data fusion \sep SAR \sep Transformer
\end{keywords}

\maketitle

\section{Introduction}\label{Introduction}

Earth observation through satellites plays a vital role in understanding the world, and has attracted attention from a wide range of communities~\citep{xia2018dota,requena2021earthnet2021,girard2021polygonal}. However, optical satellite images are often contaminated by clouds, which obstruct the view of the surface underneath, as shown in Fig.~\ref{fig:highlight-Cloudy}.
A study conducted by the MODIS instrument shows that the overall global cloudiness is roughly $67\%$ and the cloud fraction over land is about $55\%$~\citep{king2013spatial}.
Thus, cloud removal becomes an indispensable pre-processing step for applications relying on data streams of continuous monitoring~\citep{ebel2020multisensor}. 
Due to the erasure of textures in cloud-covered regions, the task of cloud removal is severely ill-posed. 
Benefiting from Synthetic Aperture Radar (SAR)~\citep{bamler2000principles} (as shown in Fig.~\ref{fig:highlight-SAR}), which is not affected by clouds due to its advantage of strong penetrability and measures the backscatter, the challenge of cloud removal can be essentially alleviated. 
However, the recovery of high-quality cloud-free images with the aid of SAR images is nevertheless a challenging problem due to the following issues:
\begin{itemize}
	\item {\bf Domain Gap}. SAR and optical images reveal different characteristics of observed objects due to their different imaging mechanisms, and thus a large domain gap exists between them~\citep{schmitt2017fusion, liu2018can}. Transferring the complementary information from a SAR image to compensate for the missing information in cloudy regions is non-trivial.
	\item {\bf Speckle Noise}. SAR images exhibit bright and dark pixels, \ie{}, speckle noise, which is uneven, even for homogeneous regions~\citep{yu2018speckle, zhu2021deep}. Moreover, the speckle noise usually exists in the same wave front as the surface information of the target. This undesirable effect leads to performance degradation on reconstruction~\citep{fuentes2019sar, liu2021mrddanet}.
\end{itemize} 

A few SAR-based cloud removal methods to learn to transfer the concatenation of multi-modal images to cloud-free images have been proposed~\citep{gao2020cloud, meraner2020cloud, ebel2020multisensor}. However, the pixel-to-pixel translation does not take into account the long-range varying contextual information of the cloud-free regions, leading to texture and structure discrepancies. Furthermore, this concatenation method only partially explores the interactions or correlations between optical and SAR data, in which complementary information cannot be effectively transferred. Moreover, simply stacking multi-modal images is susceptible to speckle noise, which hinders the cloud removal performance.

\def\sswidth{0.2\textwidth}
\begin{figure*}[!t]
	\centering
	\begin{subfigure}[b]{\sswidth}
		\centering
		\includegraphics[width=\textwidth]{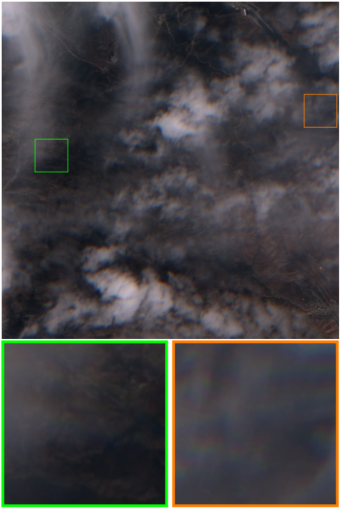}
		\caption{Cloudy}
		\label{fig:highlight-Cloudy}
	\end{subfigure}\hspace{-1mm}
	\begin{subfigure}[b]{\sswidth}
		\centering
		\includegraphics[width=\textwidth]{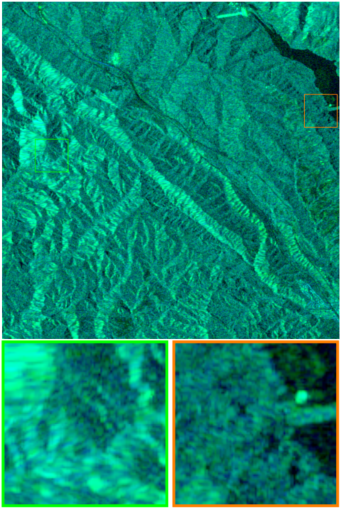}
		\caption{SAR}
		\label{fig:highlight-SAR}
	\end{subfigure}\hspace{-1mm}
	\begin{subfigure}[b]{\sswidth}
		\centering
		\includegraphics[width=\textwidth]{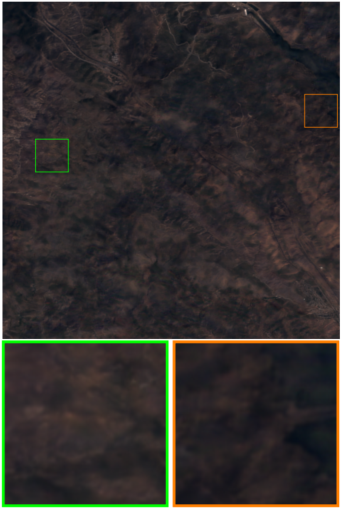}
		\caption{DSen2-CR}
		\label{fig:highlight-dsen2cr}
	\end{subfigure}\hspace{-1mm}
	\begin{subfigure}[b]{\sswidth}
		\centering
		\includegraphics[width=\textwidth]{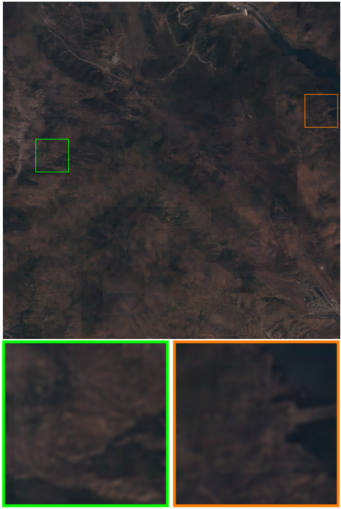}
		\caption{GLF-CR (Ours)}
		\label{fig:highlight-ours}
	\end{subfigure}\hspace{-1mm}
	\begin{subfigure}[b]{\sswidth}
		\centering
		\includegraphics[width=\textwidth]{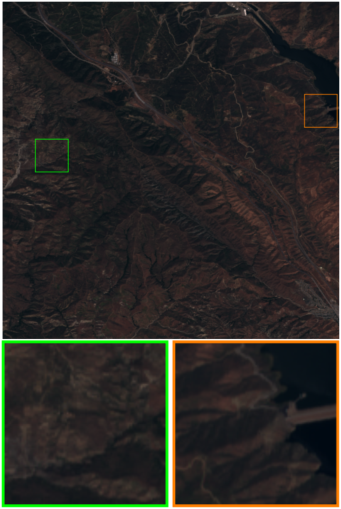}
		\caption{Cloud-Free}
		\label{fig:highlight-Cloudfree}
	\end{subfigure}
	\caption{Illustrative example of SAR-based cloud removal on a large scale cloudy image. (a) Cloudy optical image; (b) SAR image; (c) result of DSen2-CR~\citep{meraner2020cloud}; (d) result of our proposed GLF-CR; (e) cloud-free image. The GLF-CR can restore images with more details and fewer artifacts. The size of each image is $1,000 \times 1,000$ pixels.}	
	\label{fig:highlight}
	\vspace{-3mm}
\end{figure*}

To tackle the issues and limitations above, we propose a novel global-local fusion-based cloud removal (GLF-CR) algorithm by exploring the full potential of SAR image. It has been shown that SAR images help to recover texture details by compensating for the missing information in cloudy regions~\citep{meraner2020cloud}. In addition, since a SAR image is not obscured by clouds, it contains reliable global contextual information that can provide valuable guidance for capturing global interactions between contexts to maintain global consistency with the remaining cloud-free regions.
Specifically, GLF-CR contains two parallel backbones developed for optical and SAR image representation learning, where SAR features are used in a hierarchical manner to compensate for the loss of information. Inspired by Transformer architectures~\citep{vaswani2017attention} that can capture global interactions between contexts, we propose a SAR-guided global context interaction (SGCI) block in which SAR features are used to guide the interactions of global optical feature. Furthermore, a SAR-based local feature compensation (SLFC) block is proposed to transfer complementary information from the corresponding regions in the SAR features to the optical features, where dynamic filtering is used to handle speckle noise. Consequently, the proposed algorithm can generate knowledgeable features with comprehensive information, thereby yielding high-quality cloud-free images.

To sum up, the contributions of this work are three-fold:
\begin{itemize} 
	\item We propose a novel SAR-based cloud removal algorithm, {\it GLF-CR}. It incorporates the contribution of SAR to restoring reliable texture details and maintaining global consistency, thus enabling the region occluded by cloud cover to be effectively reconstructed.
	\item We propose a SAR-guided global context interaction (SGCI) block, in which the SAR feature is used to guide the global interactions between contexts in order to ensure that the structure of the recovered cloud-free region is consistent with the remaining cloud-free regions.
	\item We propose a SAR-guided local feature compensation (SLFC) block to enhance the transference of complementary information embedded in the SAR image while avoiding the influence of speckle noise, and thus generate more reliable texture details.
\end{itemize}  

\section{Related Work}
\noindent{\bf Cloud Removal.} Cloud removal aims to reconstruct the missing information caused by clouds in optical satellite imagery.
Early attempts address this problem by assuming the corrupted regions and the remaining regions share the same statistical and geometrical structures. They view cloud removal as an inpainting task and use the information around the corrupted regions to predict the missing data~\citep{chan2001nontexture, maalouf2009bandelet}. Many recent studies learn the mapping between cloudy and cloud-free images by benefiting from the remarkable generative capabilities of Generative Adversarial Networks (GANs)~\cite{singh2018cloud, wen2021generative, 9667372}. These methods fail to make accurate inferences when the corrupted region occupies a large portion of the image. 
To mitigate this problem, a series of studies make use of multispectral data to recover the missing information~\citep{shen2013compressed, xu2015thin, enomoto2017filmy}. For example, McGANs~\citep{enomoto2017filmy} and CR-GAN-PM~\citep{li2020thin} utilize additional near-infrared (NIR) images, which process higher penetrability through clouds, to improve visibility. However, as the thickness of clouds increases, all the land signals in the optical bands are obstructed. Consequently, multitemporal-based approaches have been proposed to restore the missing information with data from other time periods~\citep{scarpa2018cnn, shen2019spatiotemporal, zhang2021combined, gao2021sentinel, sen12mscrts}. However, when encountering continual cloudy days, cloud-free reference data from an adjacent period is largely unavailable.

Synthetic Aperture Radar (SAR) images are cloud-penetrable and thus provide missing information due to optically impenetrable clouds~\citep{bamler2000principles}. There is promising potential in SAR-to-optical image translation. Some researchers have tried to generate optical images directly from SAR~\citep{bermudez2018sar, fuentes2019sar}. However, since SAR lacks spectrally resolved measurements, there are domain-specific potentials and peculiarities that cannot be compensated. It is challenging to guarantee the quality of the generated optical image translated from a SAR image. Recently, a few studies have explored the means of SAR-optical data fusion, exploiting the synergistic properties of the two imaging systems to guide cloud removal.  
\cite{meraner2020cloud} concatenate the SAR image to the input optical image and use a deep residual neural network to predict the target cloud-free optical image. \cite{gao2020cloud} utilize a two-step approach, first translating the SAR image into a simulated optical image, and then concatenating the simulated optical image, the SAR image, and the optical image corrupted by clouds to reconstruct the corrupted regions using the generative adversarial network (GAN). Experiments have verified the usefulness of SAR-optical data fusion, but its gain is limited because the concatenation approach has limited ability to utilize the complementary information from the SAR image. 
To boost the gain that comes with the additional SAR information, we propose a novel cloud removal algorithm, GLF-CR, which incorporates the contribution of SAR to restoring reliable texture details and maintaining global consistency to compensate for information loss in cloudy regions.


\noindent{\bf Image Restoration.} 
Cloud removal is essentially an image restoration task in which a high-quality clean image is reconstructed from a low-quality, degraded counterpart. 
Recent advances in image restoration follow  convolutional neural network (CNN), and numerous CNN-based models have been proposed to improve restoration performance ~\citep{zhang2018image, zhang2020residual, wang2021learning}.
Global context plays an important role in local pixel-wise recovery.
However, convolution is not effective for long-range dependency modeling under the principle of local processing~\citep{liang2021swinir}.
To ensure visually consistent restoration results, a series of research focuses on the attention mechanism to obtain global dependency information.
\cite{wang2019spatial} exploit a two-round four-directional IRNN architecture to accumulate global contextual information. 
\cite{zheng2019pluralistic} introduce a short+long term attention layer to ensure appearance consistency in the image domain.
Recently, Transformer that employs a self-attention mechanism to capture global interactions between contexts~\citep{liu2021swin} has been proposed and shows promising performance in image restoration~\citep{liang2021swinir}. 
While the task of SAR-enhanced cloud removal studied in this paper needs to integrate both the information from the degraded image itself and the information from auxiliary SAR image, which is more challenging. 
 

Most existing cloud removal methods are carried out by extending the input channels of the popular CNN architectures. For example, McGAN~\citep{enomoto2017filmy} extends the input channels of the conditional Generative Adversarial Networks (cGANs) so that they are compatible with multispectral images. DSen2CR~\citep{meraner2020cloud} is derived from the EDSR network~\citep{lim2017enhanced}, and concatenates the SAR's channels and the other channels of the input optical image as input. These architectures are usually designed for tasks like super-resolution and motion deblurring, where the local information from the original low-quality image is only partially lost. For the cloud removal task, all the local information in the area covered by thick clouds is missing because the clouds completely corrupt the reflectance signal. Thus, the cloud removal methods extended from these architectures have limited ability to fully utilize the spatial consistency between the cloudy and the neighboring cloud-free regions. 
In comparison, the architecture presented in this work is designed to integrate the global context information under the guidance of the SAR image.

\noindent{\bf Multi-Modal Data Fusion.} 
Commonly used fusion strategies include element-wise multiplication/addition or concatenation between different types of features~\citep{sun2019leveraging, fu2020jl, xu2021motion}; this multi-modal data fusion yields limited performance gain~\citep{wu2018multi, audebert2018beyond, liu2021cross}.
To better exploit the complementary information of the auxiliary data, \cite{hazirbas2016fusenet} propose FuseNet for semantic segmentation. FuseNet contains two branches to extract features from the RGB and depth images, and constantly fuses them via element-wise summation. \cite{liu2021cross} propose an information aggregation distribution module for crowd counting, which consists of two branches for modality-specific representation learning (\ie{}, RGB and thermal image) and an additional branch for modality-shared representation learning. It dynamically enhances the modality-shared and modality-specific representations with a dual information propagation mechanism. These methods increase the utilization of complementary information of auxiliary data. Nevertheless, little consideration has been given to SAR-optical data fusion for cloud removal, the specific challenges of which are addressed and resolved in this work.

\section{Problem Statement}
Given a cloudy image $I$ defined over $\mathcal{X}\triangleq \mathcal{C} + \mathcal{O}$ with $\mathcal{C}$ and $\mathcal{O}$ respectively denoting the {\it cloud-covered} and {\it cloud-free regions}, the task of {\bf cloud removal} aims at restoring the cloud-covered region of the image, \ie{}, $I_{\mathcal{C}}$. Generally, this task is severely ill-posed due to the missing information caused by clouds in optical satellite observations.

\noindent{\bf Inpainting}. The basic strategy is to infer the cloud-covered region $I_{\mathcal{C}}$ from the cloud-free part $I_{\mathcal{O}}$, and thus it can be considered as {\it inpainting} task, \ie{},
\begin{equation}
	\setlength{\abovedisplayskip}{4pt}	
	\setlength{\belowdisplayskip}{4pt}
	I_{\mathcal{C}} = \mathbf{F}_\textnormal{INP}(I_{\mathcal{O}}; S(I)),
\end{equation}
where $\mathbf{F}_\textnormal{INP}$ is an inpainting operator conditioned by latent structures of the whole image, \ie{}, $S(I)$. Specifically, $S(I)$ represents priors of images, e.g., smoothness, non-local similarities, or learned features embedding from data, with which the task of cloud removal is tractable. However, latent structures of $S(I)$ are not generally holistic or are even unavailable in a cloud removal task when the cloud-covered region is dominant, leading to the failure of reconstruction if only a cloudy image $I$ is utilized.





\noindent{\bf Translation}. The SAR image $B$ is cloud free and can provide a valuable source that compensates for the information missing from the cloudy region. Inspired by the great success in style transfer work achieved by deep learning, existing SAR-based cloud removal methods mainly translate the SAR image to an optical image to remove clouds pixel-by-pixel:
\begin{equation}	
	\setlength{\abovedisplayskip}{4pt}	
	\setlength{\belowdisplayskip}{4pt}
	I_{\mathcal{C}} = \mathbf{F}_\textnormal{TRF}(B_{\mathcal{C}}; R(B,I)),
\end{equation}
where $\mathbf{F}_\textnormal{TRF}$ is a transfer operator conditioned by the inherent relationship between SAR image $B$ and optical image $I$, \ie{}, $R(B,I)$. Specifically, $R(B,I)$ represents the cross-modality transferring, which is usually learned from the dataset using the generative adversarial network (GAN) by feeding the stack of multi-modal data channels. 
However, the pixel-by-pixel translation does not take the spatial consistency between the cloudy and neighboring cloud-free regions into consideration. It consequently leads to the failure to maintain global consistency. Moreover, its method of stacking the channels of SAR and optical images is somewhat straightforward but only partially explores interactions or correlations between multi-modal data. It thus leads to limited performance improvement despite the assistance of the SAR images. And it is further influenced by the speckle noise in the SAR images, leading to reconstruction error. 

Thus the main obstacles to boosting cloud removal performance are two-fold. 
\begin{itemize}
	\item The network should effectively transfer the complementary information from SAR image $B_{\mathcal{C}}$ to the optical image while overcoming the influence of its speckle noise to generate reliable texture details.
	\item The surface information from the cloud-free region $I_\mathcal{O}$ should be considered to maintain the structure of the recovered cloud-free region consistent with the remaining cloud-free regions.
\end{itemize} 
\noindent{\bf Global-Local Fusion}. Thus the task of SAR-enhanced cloud removal is to develop an operator $\mathbf{F}_\textnormal{fusion}$ conditioned by both the inherent relationship between SAR and optical images and latent structures of the whole image, \ie{},
\begin{equation}
	\setlength{\abovedisplayskip}{4pt}	
	\setlength{\belowdisplayskip}{4pt}
	I_{\mathcal{C}} = \mathbf{F}_\textnormal{fusion}(I_\mathcal{O}, B_\mathcal{C}; S(I/B), R(B,I)),
\end{equation}
where $S(I/B)$ is the non-local context information of the cloudy image learned under the guidance of SAR image. Since the SAR image is not affected by cloud cover, it can provide valuable guidance for capturing global interactions between contexts, so as to maintain the structure of the recovered cloud-free region consistent with the remaining cloud-free regions. $R(B,I)$ in $\mathbf{F}_\textnormal{fusion}$ is different from its counterpart in $\mathbf{F}_\textnormal{TRF}$, which incorporates the information of the SAR image by stacking its channels to the optical image. We propose instead a more effective fusion strategy to transfer the complementary information from the corresponding region in the SAR image, so as to generate more reliable texture details.

\section{Method}
\subsection{Overview}
\begin{figure*}[!t]
	\centering
	\includegraphics[width=0.99\textwidth]{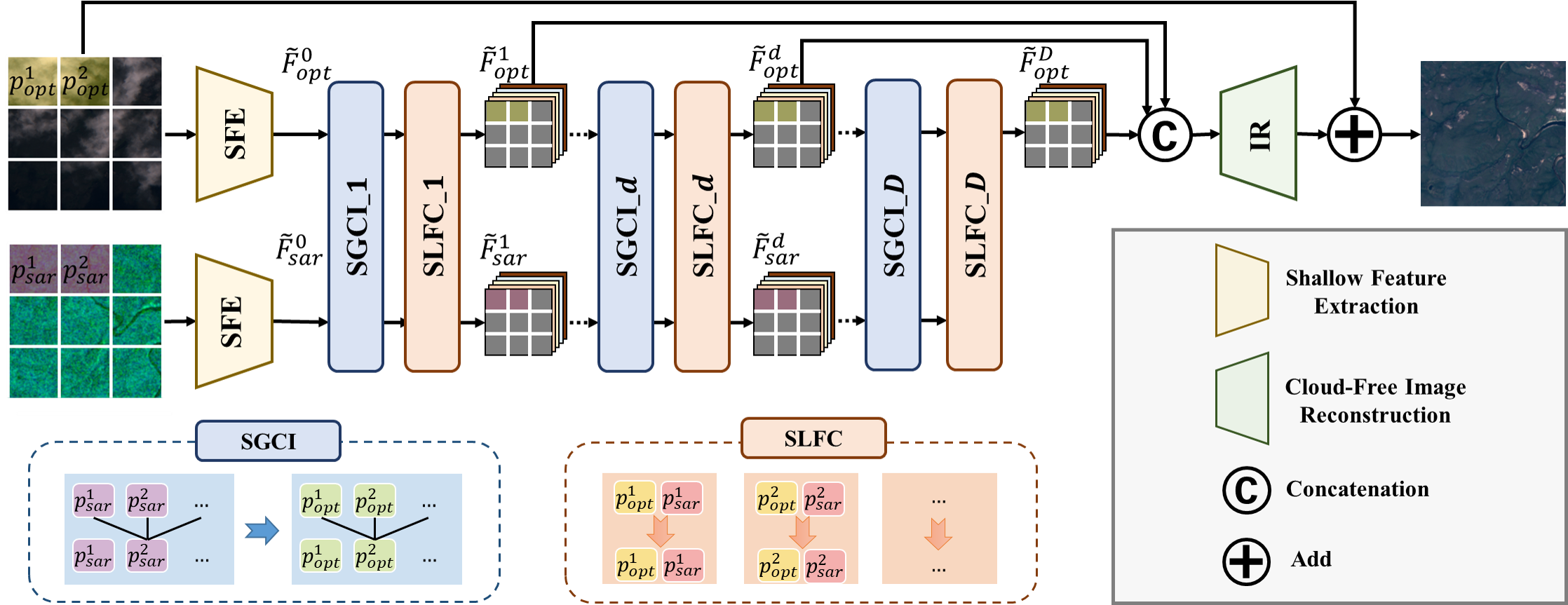}
	\caption{Overview of the proposed global-local fusion based cloud removal (GLF-CR) algorithm. It is a two-stream network in which the SAR feature is hierarchically fused into the optical feature to compensate for information loss in cloudy areas. Exploiting the power of SAR information to promote cloud removal entails two aspects: global fusion, to guide the relationship among all local optical windows based on the SAR-guided global context interaction (SGCI) block; and local fusion, to transfer the SAR feature corresponding to cloudy areas based on the SAR-based local feature compensation (SLFC) block. }
	\label{overview}
	\vspace{-5mm}
\end{figure*}

The overall framework of the proposed GLF-CR algorithm is illustrated in Fig.~\ref{overview}. It is a two-stream network in which the SAR feature is hierarchically fused into the optical feature to compensate for information loss in cloudy regions.  Exploiting the power of SAR information to promote cloud removal entails two aspects: global fusion, to guide the relationship among all local optical windows with the SGCI block; and local fusion, to transfer the SAR feature corresponding to cloudy areas with the SLFC block. Specifically, 
a cloudy image $I$ and its corresponding SAR image $B$ are first fed into different branches to extract modality-specific features $\hat{F}^0_{opt}$ and $\hat{F}^0_{sar}$ with the shallow feature extraction (SFE) block, 
\begin{equation}
	\setlength{\abovedisplayskip}{4pt}	
	\setlength{\belowdisplayskip}{4pt}
	\hat{F}^0_{opt} = H_{SFE_{opt}}(I), \hat{F}^0_{sar} = H_{SFE_{sar}}(B),
	\label{SFE}
\end{equation}
where $H_{SFE_{opt}}(\cdot)$ and $H_{SFE_{SAR}}(\cdot)$ denote the functions to extract the shallow features of the cloudy image and the SAR image, respectively.
Then, 
$\hat{F}^0_{opt}$ and $\hat{F}^0_{sar}$ are fed into $D$ functions composited from the SGCI and SLFC block to obtain knowledgeable features with comprehensive information. 
More specifically, 
the intermediate features $\{\hat{F}^1_{opt}, \hat{F}^1_{sar}\}$,  $\{\hat{F}^2_{opt}, \hat{F}^2_{sar}\}$, ..., $\{\hat{F}^D_{opt}, \hat{F}^D_{sar}\}$ are abtained as
\begin{equation}
	\setlength{\abovedisplayskip}{4pt}	
	\setlength{\belowdisplayskip}{4pt}
	\hat{F}^i_{opt}, \hat{F}^i_{sar} = H_{SLFC}(H_{SGCI}(\hat{F}^{i-1}_{opt}, \hat{F}^{i-1}_{sar})),
	\label{fusion}
\end{equation}
where $H_{SGCI}(\cdot)$ and $H_{SLFC}(\cdot)$ denote the functions of the SGCI block and the SLFC block, respectively.
The purpose of the SGCI block is local feature extraction and cross-window feature interaction, where the SAR feature is used to guide the relationship among all local optical windows. Each SGCI block is followed by an SLFC block, which is designed to fuse the complementary information from the corresponding area in a SAR image into the optical feature of a cloudy area. More details about these two blocks will be given in Secs.~\ref{Sec-SGCI} and~\ref{Sec-SLFC}.
Finally, 
the high-quality cloud-free image $I_{\mathcal{C}}$ is reconstructed by aggregating all the intermediate optical features,
\begin{equation}
	\setlength{\abovedisplayskip}{4pt}	
	\setlength{\belowdisplayskip}{4pt}
	I_{\mathcal{C}} = I + H_{IR}([\hat{F}^1_{opt}, \hat{F}^2_{opt}, ..., \hat{F}^D_{opt}]),
	\label{reconstruction}
\end{equation}
where $H_{IR}$ denotes the function of cloud-free image reconstruction, and $[\hat{F}^1_{opt}, \hat{F}^2_{opt}, ..., \hat{F}^D_{opt}]$ refers to the concatenation of the intermediate optical features. 

\subsection{SAR-Guided Global Context Interaction}
\label{Sec-SGCI}
\begin{figure}[!t]
	\centering
	\includegraphics[width=0.5\textwidth]{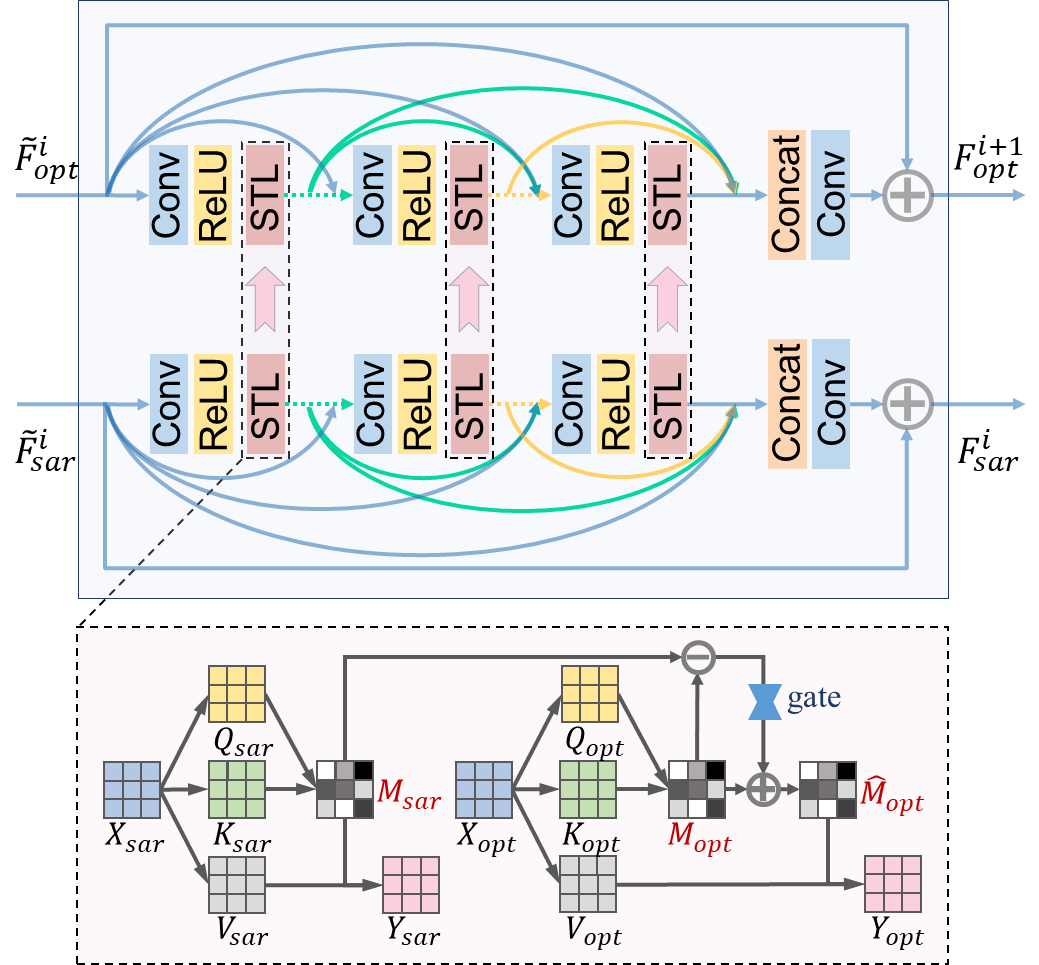}
	\caption{Detail of the SAR-guided global context interaction (SGCI) block. }
	\label{Fig-SGCI}
	\vspace{-5mm}
\end{figure}

The SGCI block, whose detail is shown in Fig.~\ref{Fig-SGCI}, has two parallel streams for the input optical and SAR features. Each stream adopts dense connections in an approach similar to the residual dense block (RDB)~\citep{zhang2018residual}, which is able to extract abundant local features via dense connected convolutional layers. As previously mentioned, SAR image clearly contributes to compensating for the missing information about cloudy regions, but not for the specific properties of optical images. Nevertheless, the cloud-free regions are conducive to the specific properties. The use of global texture information is necessary for the cloud removal task. Inspired by Transformer's ability to efficiently propagate information across the entire image to accumulate long-range varying contextual information, 
a Swin Transformer layer (STL)~\citep{liu2021swin} is added after each local convolutional layer for cross-window feature interaction.

The STL first partitions the input feature into non-overlapping $M \times M$windows, then computes the standard self-attention separately for each window. Specifically, a local window optical/SAR feature $X_{opt}/X_{sar}\in\mathbb{R}^{M^2\times C}$ is linearly transformed to query $Q_{opt}/Q_{sar}\in\mathbb{R}^{M^2\times d}$, key $K_{opt}/K_{sar}\in\mathbb{R}^{M^2\times d}$, and value $V_{opt}/V_{sar}\in\mathbb{R}^{M^2\times d}$, where $d$ is the dimension of the query or key. The attention weight matrix is computed as follows: 
\begin{equation}
	\setlength{\abovedisplayskip}{4pt}	
	\setlength{\belowdisplayskip}{4pt}
	M_{opt}\!=\!\frac{Q_{opt}K_{opt}^T}{\sqrt{d}}+B, M_{sar}\!=\!\frac{Q_{sar}K_{sar}^T}{\sqrt{d}}+B,
	\label{self-attn}
\end{equation}
where $B$ is the learnable relative positional encoding. The essence of this attention matrix is the weight of a particular region that is absorbing information from other regions. For a cloudy region, due to the information loss, it is difficult to estimate its interactions with cloud-free regions. For the same region in a SAR image, its interactions with other regions can be estimated easily, which provides valuable guidance for the interactions between cloud-free and cloudy regions in the optical image. Thus, 
we transfer the attention map of the SAR image to refine the attention map of the optical image,
\ie{}, 
we use $M_{sar}$ to improve $M_{opt}$. 
We first obtain the attention map of the optical and SAR features $M_{opt}$ and $M_{sar}$ by Eq.~(\ref{self-attn}).
Then we compute the difference between $M_{opt}$ and $M_{sar}$ and obtain $M_{res}$. Finally, we apply a gating function to adaptively refine $M_{opt}$:
\begin{equation}
	\setlength{\abovedisplayskip}{4pt}	
	\setlength{\belowdisplayskip}{4pt}
	\hat{M}_{opt} = M_{opt} + M_{res}\odot G(M_{res}),
	\label{self-attn-opt-update}
\end{equation}
where $G(\cdot)$ is the gating function fed with the residual term $M_{res}$ and $\odot$ denotes the  element-wise multiplication operation. The optical and SAR output are computed as:
\begin{equation}
	\setlength{\abovedisplayskip}{4pt}	
	\setlength{\belowdisplayskip}{4pt}
	Y_{opt}\!=\!\mbox{Softmax}(\hat{M}_{opt})V_{opt},
	Y_{sar}\!=\!\mbox{Softmax}(M_{sar})V_{sar}.
	\label{self-attn-opt-output}
\end{equation}
This module considers the relationship among all local window optical features under the guidance of the SAR feature, denoted in this paper as global fusion.

\subsection{SAR-based Local Feature Compensation}
\label{Sec-SLFC}
\begin{figure}[!t]
	\centering
	\includegraphics[width=0.48\textwidth]{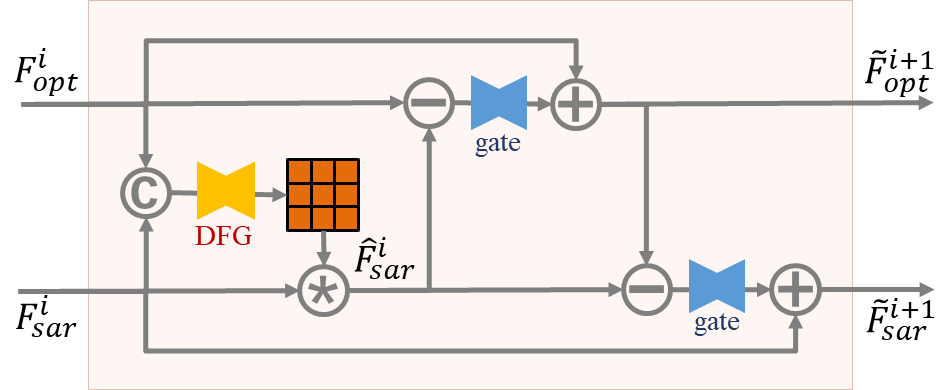}
	\caption{Detail of the SAR-based local feature compensation (SLFC) block. }
	\label{Fig-SLFC}
	\vspace{-5mm}
\end{figure}

The detail of the SLFC block is shown in Fig.~\ref{Fig-SLFC}.
Because the SAR image is corrupted by severe speckle noise, we utilize dynamic filtering for SAR features before information transference.  Standard convolution filters are shared across all pixels in an image, while the dynamic filters vary
from pixel to pixel. Therefore, the dynamic filters can handle the spatial variance issue~\citep{jia2016dynamic, zhou2019spatio}, thus helping to suppress the spatially unevenly speckle noise. Specifically, 
a filter is dynamically generated for each position in the SAR feature using the Dynamic Filter Generation (DFG) module.
The DFG module takes the concatenation of the optical and SAR features $Concat(F^i_{opt}, F^i_{sar})\in\mathbb{R}^{H\times W\times 2C}$ as input. The dimension of the generated filter $\mathcal{F}^i$ is $H\times W \times Ck^2$ and is reshaped into a five-dimensional filter. Then, for each position
$(h, w, c)$ in the SAR feature $F^i_{sar}\in\mathbb{R}^{H\times W\times C}$, a specific local filter
$\mathcal{F}^i(h, w, c)\in\mathbb{R}^{k\times k}$ is applied to the region centered around $F^i_{sar}(h, w, c)$ as
\begin{equation}
	\setlength{\abovedisplayskip}{4pt}	
	\setlength{\belowdisplayskip}{4pt}
	\hat{F}^i_{sar}(h, w, c)= \mathcal{F}^i(h, w, c)\ast F^i_{sar}(h, w, c),
	\label{dynamic filter}
\end{equation}
where $\ast$ denotes the convolution operation. 


After transforming the extracted SAR feature $ F^i_{sar}$ using the dynamic filter to improve tolerance of speckle noise, we propagate the complementary information from the SAR feature to refine the optical feature, in the same way that the attention map is refined. We compute the difference between the optical and SAR features to obtain the residual information $F^i_{s-o}=\hat{F}^i_{sar}-F^i_{opt}$, and apply a gating function to transfer the complementary information,
\begin{equation}
	\setlength{\abovedisplayskip}{4pt}	
	\setlength{\belowdisplayskip}{4pt}
	\tilde{F}^i_{opt} = {F}^i_{opt} + F^i_{s-o}\odot G(F^i_{s-o}).
\end{equation}
To better exploit interactions among elements of the optical and SAR features for a further performance gain, we adopt a dual information propagation
mechanism, \ie{}, updating the SAR feature as well. We compute the difference between the SAR feature and the updated optical feature $F^i_{o-s}=\tilde{F}^i_{opt}-\hat{F}^i_{sar}$, and also propagate the information through use of a gating function,
\begin{equation}
	\setlength{\abovedisplayskip}{4pt}	
	\setlength{\belowdisplayskip}{4pt}
	\tilde{F}^i_{sar} = {F}^i_{sar} + F^i_{o-s}\odot G(F^i_{o-s}).
\end{equation}
The enhanced optical and SAR features are then introduced into the next SGCI for further representation learning. This module considers the information transference between local features, denoted as local fusion in this paper.

\section{Experiments}
\subsection{Experimental Settings}
\noindent{\bf Dataset and Metrics}.
The experiments are conducted on the large-scale dataset SEN12MS-CR~\citep{ebel2020multisensor}, which is built from freely available data acquired by the Sentinel satellites in the Copernicus program. The dataset contains $122,218$ samples from 169 non-overlapping regions of interest (ROI) distributed over all inhabited continents during all meteorological seasons. Each sample consists of a triplet of an orthorectified, geo-referenced Sentinel-1 dual-pol SAR image, a Sentinel-2 cloud-free multi-spectral image, and a cloud-covered Sentinel-2 multi-spectral image where the observations of cloud-free and cloud-covered images are close in time. The size of each image is $256 \times 256$ pixels. 
The VV and VH polarizations of the SAR images are clipped to values $[-25, 0]$ and $[-32.5, 0]$, and rescaled to the range $[0, 1]$. All bands of the optical images are clipped to values $[0, 10000]$, and rescaled to the range $[0, 1]$ as well. 
We split the 169 ROIS into 149 ROIs for training, 10 ROIs for validation, and 10 ROIs for test. To avoid overall performance being biased towards specific cloud cover level, we calculate the percentage of cloud cover of each image by utilizing the cloud detection flowchart in \cite{meraner2020cloud} and randomly select 800 samples from the samples with cloud cover of $0\%$ to $20\%$, $20\%$ to $40\%$, $40\%$ to $60\%$, $60\%$ to $80\%$, and $80\%$ to $100\%$ as the test set, respectively. Specifically, the training, validation and test set consist of $101,615$, $8,623$ and $4,000$ samples respectively. 
The results of cloud removal are evaluated with the normalized data based on the peak signal-to-noise ratio (PSNR), structural similarity index measure (SSIM), spectral angle mapper (SAM), and mean absolute error (MAE).

\noindent{\bf Implementation Details}. 
The proposed GLF-CR network is implemented using publicly available Pytorch and trained in an end-to-end manner supervised by L1 loss on 4 NVIDIA TITAN V GPUs. We implement the gating functions in Secs.~\ref{Sec-SGCI} and~\ref{Sec-SLFC} by employing a convolution layer as well as a Softmax layer, and the dynamic filter generation (DFG) module in Sec.~\ref{Sec-SLFC} is constituted by a convolution layer followed by two residual blocks. 
During training, we randomly crop the samples into $128 \times 128$ patches. In an empirical manner, the batch size is set to 12 and the maximum epoch of training iterations is set to 30. The Adam optimizer is used and the learning rate starts at $10^{-4}$, which decays by $50\%$ every five epochs. By trading off the performance and complexity of the model, the number of the SGCI and SLFC blocks $D$ is set to $6$; the number of dense connections in each stream of the SGCI block is set to $5$; the window size and the attention head number for the STL layer are set to $8$ and $8$, respectively; and the size of the dynamic filter $k$ is set to $5$. The codes, models, and more results are released at: \url{https://github.com/xufangchn/GLF-CR}.

\subsection{Comparisons with State-of-the-art Methods}
We compare the proposed GLF-CR networks to state-of-the-art cloud removal methods, including multi-spectral based approaches, 
SpA GAN~\citep{pan2020cloud},  
the SAR-to-optical image translation approach, 
SAR2OPT~\citep{bermudez2018sar}, 
and SAR-optical data fusion based approaches,
SAR-Opt-cGAN~\citep{grohnfeldt2018conditional}, Simulation-Fusion GAN~\citep{gao2020cloud} and DSen2-CR~\citep{meraner2020cloud}.
SpA GAN takes all channels of the input optical image as input. It uses the spatial attention network (SPANet)~\citep{wang2019spatial} as a generator to model the map from a cloudy image to a cloudless image. 
SAR2OPT performs SAR-to-optical translation by takeing the U-Net as the generator, not relying on any (cloudy) optical satellite information.
SAR-Opt-cGAN and DSen2-CR both leverage the SAR image as a form of prior to guide the reconstruction process under thick, optically impenetrable clouds. The SAR's channels are simply concatenated to the other channels of the input optical image to predict the full spectrum of optical bands. SAR-Opt-cGAN is extended from U-Net, while DSen2-CR is extended from the EDSR network~\citep{lim2017enhanced}.
Simulation-Fusion GAN first translates the SAR image into simulated optical data, then takes the concatenation of the simulated optical image, SAR and the corrupted optical image as input for prediction.

To validate the superiority of GLF-CR in leveraging the power of SAR images, we also refer to the fusion strategy in SAR-Opt-cGAN and DSen2-CR to train the proposed network, by using concatenation, denoted as {\it Concat}. We concatenate the SAR's channels and optical image's channels as input, and remove the branch for SAR feature learning, the attention map update in the SGCI blocks, and the SLFC blocks. The quantitative results are presented in Table~\ref{sota}. The proposed GLF-CR network brings remarkable improvements compared to state-of-the-art methods. We choose 3 scenes to evaluate qualitative results, as shown in Fig.~\ref{fig:SOTA}. For each scene, from top-left to bottom-right are respectively the cloudy image, the SAR image, 
the results from SpA GAN,  SAR2OPT, SAR-Opt-cGAN, Simulation-Fusion GAN, DSen2-CR, {\it Concat} and GLF-CR, 
and the cloud-free image. We find that the proposed GLF-CR network achieves the best visualization performance. Detailed analyses are presented below.

\begin{table*}[!t]
	\small
	\centering
	\caption{Quantitative comparisons of proposed GLF-Nets to state-of-the-art methods.}
	\label{sota}
	\resizebox{.98\textwidth}{!}{
		\begin{tabular}{ccccccccc}
		\hline
		\multirow{2}{*}{Method}&  & \multicolumn{2}{c}{Input}&  & \multirow{2}{*}{PSNR $\uparrow$} & \multirow{2}{*}{SSIM $\uparrow$} & \multirow{2}{*}{SAM $\downarrow$} & \multirow{2}{*}{MAE $\downarrow$} \\ \cline{3-4}
         &  & \multicolumn{1}{c}{Optical} & \multicolumn{1}{c}{SAR} & & & & & \\ \hline
         SpA GAN~\citep{pan2020cloud}           & & \checkmark & \xmark     &  & $24.8688$    & $0.7533$    & $16.0454$    & $0.0444$     \\ \hline 
         SAR2OPT~\citep{bermudez2018sar}        & & \xmark     & \checkmark &  & $25.7223$    & $0.7918$    & $14.0501$    & $0.0427$     \\ \hline 
         SAR-Opt-cGAN~\citep{grohnfeldt2018conditional}&&\checkmark&\checkmark&& $25.2948$    & $0.7594$    & $14.4389$    & $0.0441$     \\ 
         Simulation-Fusion GAN~\citep{gao2020cloud} & & \checkmark & \checkmark     &  & $24.5519$    & $0.6947$    & $15.5929$    & $0.0455$     \\ 
         DSen2-CR~\citep{meraner2020cloud}      & & \checkmark & \checkmark &  & $27.3780$    & $0.8705$    & $8.5073$    & $0.0319$     \\
         Concat (Ours)                          & & \checkmark & \checkmark &  & $28.5324$    & $0.8804$    & $8.1088$    & $0.0284$     \\
         GLF-CR (Ours)                          & & \checkmark & \checkmark &  &$\bm{29.0793}$&$\bm{0.8855}$&$\bm{7.6455}$&$\bm{0.0266}$ \\ \hline 
        \end{tabular}
	}
	\vspace{-4mm}
\end{table*}

\input{fig_sota}

%
We first compare the cloud removal performance of SAR-based methods to the conventional method, SpA GAN. As the SAR image encodes rich geometrical information about cloud-covered regions, it facilitates the ground object construction. SpA GAN, which relies solely on cloudy optical images, are less effective than SAR-based cloud removal methods.
As shown in Fig.~\ref{fig:SOTA},
it fails to tackle the thick cloud removal and generates undesirable artifacts, especially for cloud-covered regions.

We next compare the cloud removal performance of the SAR-to-optical image translation approach, SAR2OPT to the SAR-optical data fusion based approaches. SAR2OPT, which relies solely on SAR images, can reconstruct prominent geometric characteristics related to roads, crop fields, etc. But it suffers from content vanishing because the specific potentials and peculiarities of optical images cannot be fully compensated from the SAR images. Moreover, a distinct difference in the color distribution of SAR2OPT's reconstruction results and ground truth can be observed. SAR-Opt-cGAN adopts the same generator architecture as SAR2OPT while taking both the cloudy optical image and the SAR image as input. 
However, it performs worse than SAR2OPT which only takes the SAR images as input. 
And as shown in the second scene of Fig.~\ref{fig:SOTA}, the SAR image clearly emphasizes the surface's physical properties. SAR-Opt-cGAN fails to reconstruct it while SAR2OPT does. It demonstrates the challenge of taking advantage of multi-modal data fusion while avoiding the performance degradation caused by the undesirable effects in each modality.
Simulation-Fusion GAN suffers from the performance degradation caused by the undesirable effects in simulated optical image besides the cloudy optical and SAR images, and also has poor color fidelity. 
To some extent, DSen2-CR alleviates the performance degradation caused by the undesirable effects by utilizing a tailored generator. However, its gain is still limited.

Our methods perform favorably when compared with DSen2-CR, which exploits the inherent advantage of SAR image. Among them, {\it Concat} adopts the same approach as SAR-Opt-cGAN and DSen2-CR to utilize the complementary information embedded in SAR images. It achieves higher performance than SAR-Opt-cGAN and DSen2-CR, as shown in Table~\ref{sota}. Unlike the approach of SAR-Opt-cGAN and DSen2-CR, {\it Concat} contains global context interactions, which takes the information embedded in neighboring cloud-free regions into consideration, thus performing better in terms of global consistent structure. But those methods still leave distinct clouds or blur some image textures, which reflects the limitations of the concatenation method. Furthermore, It can be observed that the proposed GLF-CR network outperforms other methods by a large margin. It can restore images with more details and fewer artifacts, as shown in Fig.~\ref{fig:SOTA}. These significant improvements demonstrate that the proposed method can better use the complementary information embedded in SAR images.

\subsection{Analysis on Different Cloud Cover Levels}
\begin{figure*}[!t]
     \centering
     \includegraphics[width=0.98\textwidth]{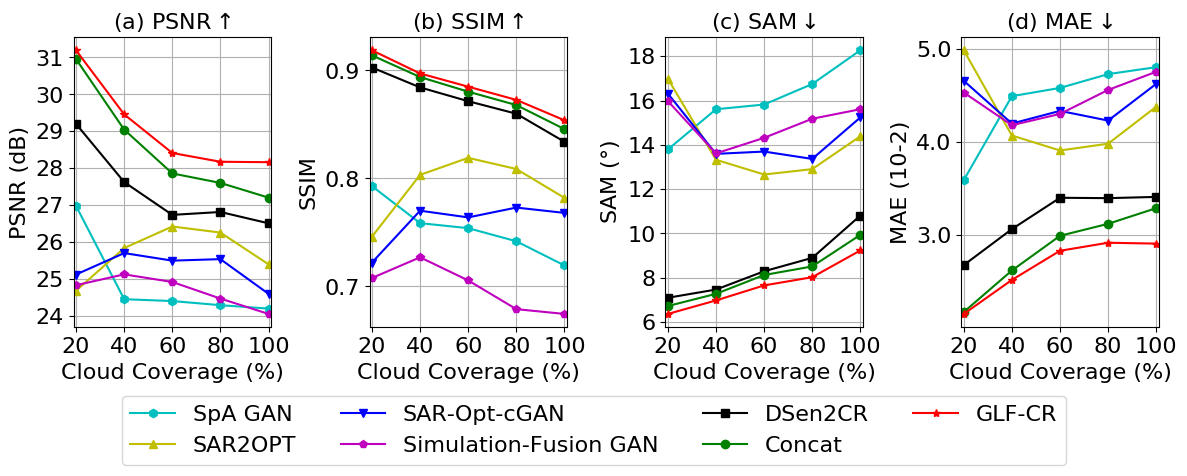}
     \vspace{-2mm}
     \caption{Quantitative comparisons of proposed GLF-Nets to state-of-the-art methods on different cloud cover levels in terms of the PSNR, SSIM, SAM, and MAE quality metrics.}
     \label{fig:cloud-cover}
     \vspace{-7mm}
\end{figure*}

We further compare the proposed GLF-CR networks to state-of-the-art cloud removal methods on different cloud cover levels. We evaluate the performance of cloud removal on the images with cloud cover of $0\%$ to $20\%$, $20\%$ to $40\%$, $40\%$ to $60\%$, $60\%$ to $80\%$, and $80\%$ to $100\%$, and show the comparison results in terms of the PSNR, SSIM, SAM, and MAE quality metrics in Fig.~\ref{fig:cloud-cover}. The proposed methods perform favorably when compared with state-of-the-art methods on all cloud cover levels.

It is observed that the overall performance of multispectral-based approaches, SpA GAN, is negatively correlated with the cloud cover level. With the higher cloud cover level, they get less prior information and thus perform worse. And the performance of the SAR-to-optical image translation approach, SAR2OPT, is not related to the cloud cover level. 

SAR-Opt-cGAN and Simulation-Fusion GAN utilize the prior information from both cloudy images and SAR images. It suffers the performance degradation caused by the undesirable effects in both modalities. When the cloud cover is low, it is not as good as the multispectral-based methods due to the interference from additional SAR image or simulated optical image from SAR image. When the cloud cover is high, it is not as good as the SAR-to-optical image translation approach due to the interference from clouds.

DSen2-CR alleviates the performance degradation to some extent, and thus outperforms the single-modality-based methods. {\it Concat} adopts the same fusion strategy in DSen2-CR to utilize the complementary information embedded in SAR images while takes the information embedded in neighboring cloud-free regions into consideration, thus its performance is more superior to that of DSen2-CR when more prior information from cloud-free regions is available.
And the proposed method is superior in exploiting the power of SAR information in addition to considering the information embedded in neighboring cloud-free regions, and thus steadily outperforms DSen2-CR on all cloud cover levels.

%

\subsection{Ablation Study}
\begin{figure*}[!t]
     \centering
     \includegraphics[width=0.98\textwidth]{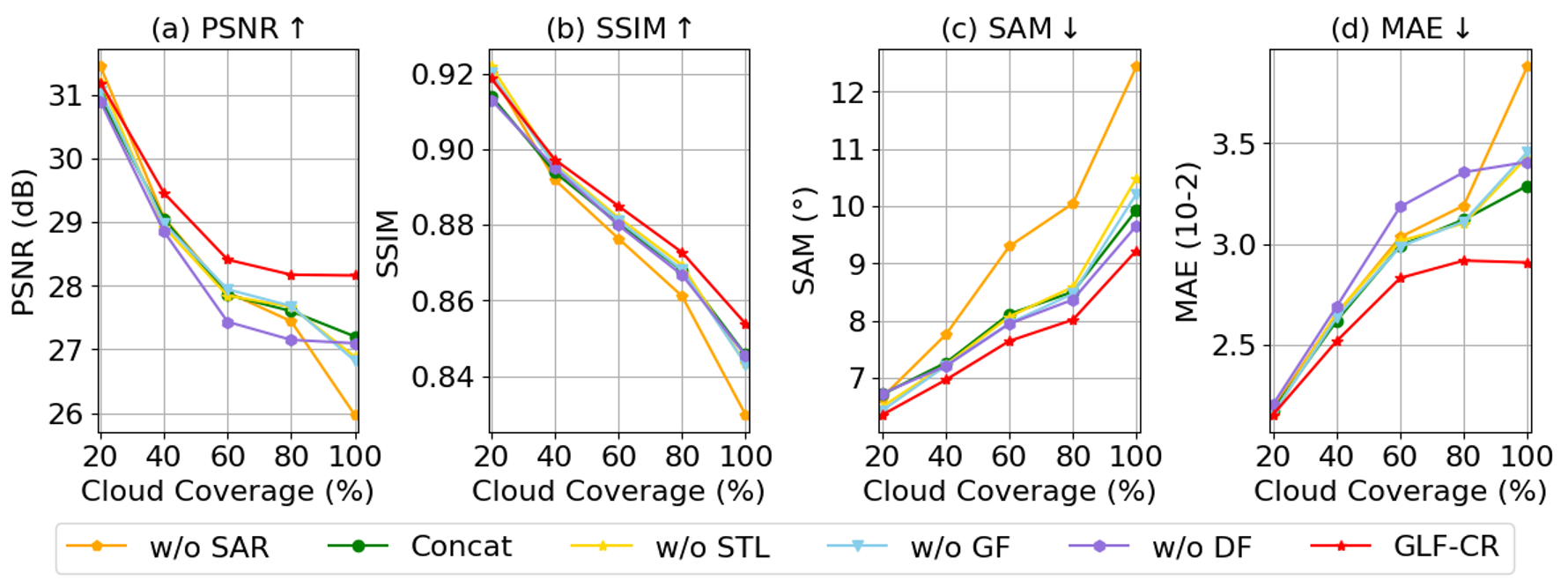}
     \vspace{-3mm}
     \caption{Quantitative ablation study on different cloud cover levels in terms of the PSNR, SSIM, SAM and MAE quality metrics.}
     \label{fig:ablation_cloud}
     \vspace{-5mm}
\end{figure*}

The proposed GLF-CR network improves the performance of SAR-based cloud removal by incorporating global fusion to guide the relationship among all local optical windows with SAR features and local fusion to transfer the SAR feature corresponding to cloudy areas to compensate for the missing information. To determine what contributes to the superior performance of the proposed approach, we analyze the effectiveness of each component by comparing a few variants with and without the use of SAR image (SAR), Concatenation fusion (Concat), STL layer (STL), global fusion (GF), and dynamic filter (DF). The qualitative and qualitative results are shown in Table~\ref{ablation-study-table} and Fig.~\ref{ablation-study-figure}, and the results on different cloud cover levels is shown in Table~\ref{fig:ablation_cloud}.
From the table and the figure, we can draw the following conclusions:

\begin{table}[!t]
	\small
	\centering
	\caption{Quantitative ablation study of proposed algorithm with and without use of the SAR image (SAR), Concatenation fusion (Concat), STL layer (STL), global fusion (GF), and dynamic filter (DF).}
	\label{ablation-study-table}
	\resizebox{.49\textwidth}{!}{
		\begin{tabular}{c|cccc}
			\hline
			Method & PSNR $\uparrow$& SSIM $\uparrow$& SAM $\downarrow$& MAE $\downarrow$\\ \hline
			w/o SAR&  $28.3657$       &  $0.8759$        &  $8.1783$         &  $0.0299$         \\ 
			Concat &  $28.5324$       &  $0.8804$        &  $8.1088$         &  $0.0284$         \\ 
			w/o STL&  $28.5079$       &  $0.8825$        &  $8.1783$         &  $0.0287$         \\ 
			w/o GF &  $28.4983$       &  $0.8816$        &  $8.0595$         &  $0.0287$         \\ 
			w/o DF &  $28.2867$       &  $0.8800$        &  $7.9853$         &  $0.0297$         \\ 
			GLF-CR &  $\bm{29.0793}$  &  $\bm{0.8855}$   &  $\bm{7.6455}$    &  $\bm{0.0266}$    \\ \hline 
		\end{tabular}
	}
	\vspace{-5mm}
\end{table}

\def\sswidth{0.98\textwidth}
\begin{figure*}[!h]
    \centering
    \begin{subfigure}[b]{\sswidth}
     \centering
     \includegraphics[width=\textwidth]{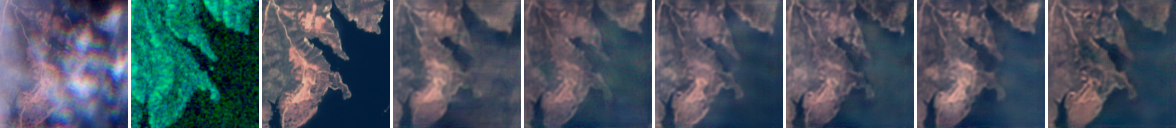}
    \end{subfigure}
    \\
    \begin{subfigure}[b]{\sswidth}
     \centering
     \includegraphics[width=\textwidth]{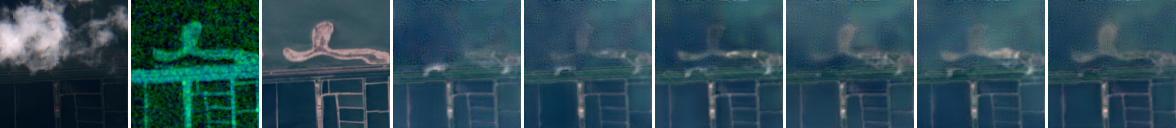}
    \end{subfigure}
    \\
    \begin{subfigure}[b]{\sswidth}
     \centering
     \includegraphics[width=\textwidth]{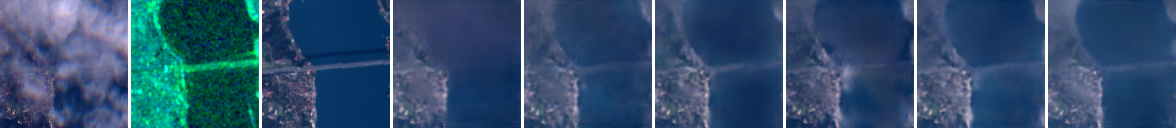}
    \end{subfigure}
    \\
    \begin{subfigure}[b]{\sswidth}
     \centering
     \includegraphics[width=\textwidth]{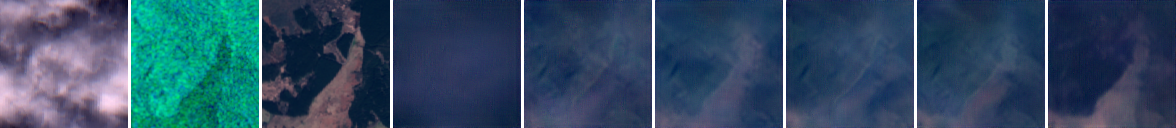}
    \end{subfigure}
    \\
    \begin{minipage}{\sswidth}
    \begin{tabu} to \textwidth {X[c] X[c] X[c] X[c] X[c] X[c] X[c] X[c] X[c]}
    Cloudy & SAR & Cloud-Free & w/o SAR & Concat & w/o STL & w/o GF & w/o DF & GLF-CR
    \end{tabu}
    \end{minipage}
    \caption{Qualitative ablation study with 4 scenes by different GLF-CR networks. For each scene, from left to right, are respectively the cloudy image, the SAR image, the cloud-free image, and the result by {\it w/o SAR}, {\it Concat}, {\it w/o STL}, {\it w/o GF}, {\it w/o DF}, and GLF-CR. The size of each image is $128 \times 128$ pixels.}
    \label{ablation-study-figure}
    \vspace{-5mm}
\end{figure*}

\noindent{\bf Importance of SAR Image}. 
We validate the importance of the SAR image by training the GLF-CR network without SAR images, 
denoted as {\it w/o SAR}. Since the input is a single source signal, \ie{}, the cloudy optical image itself, a single-stream network is adopted and no fusion strategy is used. 
As shown in Fig.~\ref{fig:ablation_cloud}, it performs comparable to the networks employing SAR images when the cloud cover level is low. However, when the cloud cover level gets higher, the performance gap between the networks with and without SAR images gets larger.
And as shown in Fig.~\ref{ablation-study-figure}, {\it w/o SAR} tends to generate over-smoothed effects for cloud-covered regions, while the networks with SAR images can recover texture details. This demonstrates that the rich complementary information encoded in SAR images can effectively improve the cloud removal performance. 

\noindent{\bf Limitation of Concatenation Fusion}.  Compared with {\it w/o SAR}, {\it Concat} only adds two channels to the input to utilize the SAR image. 
The gain of utilizing the concatenation fusion is 0.17dB, while the proposed GLF-CR network obtains a gain of 0.71dB. As observed from Fig.~\ref{fig:ablation_cloud}, when the proportion of cloud-free regions is higher, the performance gap between {\it Concat} and {\it GLF-CR} is larger, since the proposed GLF-CR network can better exploit the power of SAR information compared with the concatenation fusion.
Fig.~\ref{ablation-study-figure} shows that the proposed GLF-CR network can recover more complete texture structure and obtain better visual effects.

\noindent{\bf Effectiveness of Global Interactions}. Capturing global interactions between contexts plays a vital role in maintaining global consistent structure. 
We train the GLF-CR network by removing the STL layers in the SGCI blocks, denoted as {\it w/o STL}. 
It can be observed that the proposed GLF-CR method which captures the global interactions between contexts can improve cloud removal performance effectively. 
It recovers clearer and more complete structure for the land in the second and fourth scenes in Fig.~\ref{ablation-study-figure}.

\noindent{\bf Effectiveness of SAR-Guided Global Interactions}. We further validate the effectiveness of guiding the global interactions of optical features with SAR features.
We train the GLF-CR network by reserving the STL layer but not using the SAR feature to guide the global optical interactions, denoted as {\it w/o GF}.
Compared with {\it w/o STL}, it can be observed that {\it w/o GF} has only a slight performance improvement in terms of SAM, despite using additional STL layers to maintain the spatial consistency, since estimating the interactions from the cloudy optical image itself will introduce some error. As shown in the third scene in Fig.~\ref{ablation-study-figure}, {\it w/o GF} generates undesirable artifacts. As the SAR image is not affected by cloud cover, it can provide valuable guidance for capturing global interactions between contexts. 
This point can be validated by comparing the results of {\it w/o GF} and {\it GLF-CR}. 
It can be seen that guiding the global interactions of optical features with SAR features can effectively improve the performance of cloud removal and make the structure of the predicted cloud-free image more consistent with ground truth.

\noindent{\bf Effectiveness of the Dynamic Filter}. The proposed GLF-CR network uses dynamic filtering to handle the speckle noise of SAR images. To validate the effectiveness of the dynamic filter, we train the GLF-CR network by removing the dynamic filter in SLFC blocks, denoted as {\it w/o DF}. 
It can be seen from Fig.~\ref{fig:ablation_cloud} that the performance of {\it w/o DF} degrades more severely in terms of PSNR and MAE that measure the quality of reconstructed images than in terms of SSIM and SAM that quantify spectral and structural similarity. And it can be observed that the trends of {\it w/o DF} and {\it GLF-CR} relative to the cloud cover level are similar. As both methods adopt the same strategy to utilize the information of the cloud-free regions and SAR images, while the proposed GLF-CR network can alleviate the problem of speckle noise in the SAR image and generate clearer images.

\section{Discussion}
\begin{figure}[!t]
    \centering
    \begin{subfigure}[b]{0.48\textwidth}
     \centering
     \includegraphics[width=\textwidth]{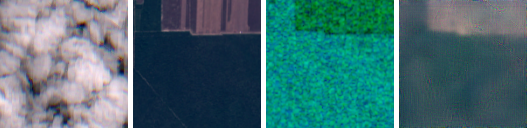}
    \end{subfigure}
    \\
    \begin{subfigure}[b]{0.48\textwidth}
     \centering
     \includegraphics[width=\textwidth]{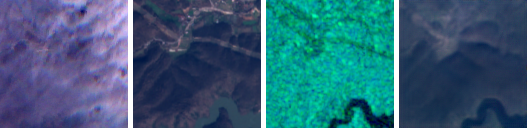}
    \end{subfigure}
    \\
    \begin{subfigure}[b]{0.48\textwidth}
     \centering
     \includegraphics[width=\textwidth]{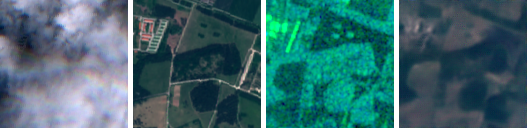}
    \end{subfigure}
    \\
    \begin{subfigure}[b]{0.48\textwidth}
     \centering
     \includegraphics[width=\textwidth]{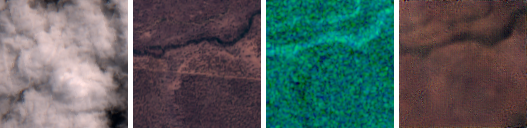}
    \end{subfigure}
    \\
    \begin{minipage}{0.48\textwidth}
    \begin{tabu} to \textwidth {X[c] X[c] X[c] X[c]}
    Cloudy & Cloud-Free & SAR & GLF-CR
    \end{tabu}
    \end{minipage}
    \caption{Example results of GLF-CR on the images completely obscured by clouds.}
    \label{fig:challenge}
    \vspace{-5mm}
\end{figure}

\noindent{\bf Performance on Challenging Conditions}.
Cloud removal is quite challenging when the image to be processed is completely cloudy. To see how the proposed method behaves in the challenging conditions, Fig.~\ref{fig:challenge} shows the results on the images where the ground information is almost obscured by clouds. It can be found that the proposed method can recover the approximate information of ground objects while with poor texture details. Since the images are completely cloudy, no cloud-free part can be accessed and only SAR information is available to reconstruct the cloud-free images. The quality of reconstructed cloud-free images depends entirely on the information embedded in the SAR image. While the SAR image fails to feature the different agricultural landscapes, as seen in the first scene in Fig.~\ref{fig:challenge}, the reconstructed cloud-free image loses the corresponding details. And since no spectral information is available, the spectral fidelity of the  reconstructed cloud-free image degrades.

\noindent{\bf Speckle Noise in SAR Data}.
The SAR data in SEN12MS-CR dataset is from the Level-1 GRD product, which has been multi-looked for reduced speckle. Notwithstanding, the multi-looked data still exhibits a high degree of speckle noise, as seen from Fig.~\ref{fig:highlight}, \ref{fig:SOTA} and \ref{ablation-study-figure}, since speckle noise is multiplicative in nature and difficult to distinguish from the original signal. And, while commonly referred to as ``speckle noise'', speckle is not only noise but in some sense has an information content~\citep{argenti2013tutorial}. At this point, we do not consider an explicit despeckling preprocessing step, but implicitly handle the spatially varying speckle distribution by the dynamic filter embedded in the network. It is also possible to preprocess the SAR data with a despeckling technique before feeding it to the network. Therefore, we train the GLF-CR network by removing the dynamic filter in SLFC blocks while feeding the SAR data despeckled with a median filter. As shown in Table.~\ref{despeckled-table}, we can see that preprocessing the SAR data with a despeckling technique can reduce the influence of speckle noise on cloud removal. While the proposed method implicitly mitigates the influence of speckle noise based on the dynamic filter and can achieve better performance.  

\noindent{\bf Geometric Distortion in SAR Data}.
It is well-known that there is an inherent geometric distortion in SAR data when the terrain is undulating, due to the sensor's sideways view. It will lead to the inconsistency between the information in the SAR data and the actual state of the ground objects, adversely affecting the cloud removal performance. The experiments in this paper are conducted on the SEN12MS-CR dataset (to our best knowledge, the only open-source cloud removal dataset with SAR data), where the SAR data is provided by the Sentinel-1 satellites. Its resolution is $10m$ and thus does not show excessive distortion. Furthermore, depending on the large scale of the dataset, the proposed powerful model can address this aspect to some extent. 

\begin{table}[!t]
    \small
    \centering
    \caption{Performance of proposed algorithm with use of despeckled SAR data.}
    \label{despeckled-table}
    \resizebox{.49\textwidth}{!}{  
    	\begin{tabular}{c|cccc}
    		\hline
    		Method & PSNR $\uparrow$& SSIM $\uparrow$& SAM $\downarrow$& MAE $\downarrow$\\ \hline
    		w/ despeckled SAR &  $28.5377$       &  $0.8818$        &  $8.0719$         &  $0.0286$\\
    		w/o DF &  $28.2867$       &  $0.8800$        &  $7.9853$         &  $0.0297$         \\
    		GLF-CR &  $ 29.0793 $  &  $ 0.8855 $   &  $7.6455$    &  $0.0266$    \\ \hline 
    	\end{tabular}
    }
    \vspace{-5mm}
\end{table}

\noindent{\bf Registration error between the optical and SAR Data}.
The registration error between the optical image and its corresponding SAR image is expected to affect the learning process. The data instructions given by ESA illustrate that the Sentinel-1 SAR L1 productions and the Sentinel-2 optical L1C productions have a co-registration accuracy of within 2 pixels. We set the size of the dynamic filter in the SLFC blocks to $5$ for a larger receptive field, which allows the proposed model to work when tiny deviations exist between the SAR and optical images. 

\noindent{\bf Nuisances between Cloudy Reference Image and Cloud-Free Target Image}.  The cloud removal performance in the paper is assessed on the SEN12MS-CR dataset by comparing the prediction with the cloud-free image temporally close to the cloudy one. There are some inevitable nuisances determined by the sunlight condition, acquisition geometry, humidity, pollution, change of landscape, etc, while the SEN12MS-CR dataset is curated to minimize such cases. However, the inevitable nuisances are negligible for a relatively large-scale test split that is globally and seasonally sampled without any bias to specific sunlight condition, etc. It implies that models biased to specific condition won't have any unfair advantages on the test split. Overall, the influence of nuisances can be averaged out. It poses no concern about the fairness of benchmarking the proposed model on the considered dataset. 

In addition, we test the proposed method on images where the interval between the cloud-free and cloudy image is different. The date of input cloudy image is July 17, 2018, and we use the SAR image with the closest interval to cloudy image as auxiliary data, whose date is July 18, 2018. And the date of cloud-free images used for the assessment are July 30, 2018, August 11, 2018 and September 28, 2018, respectively. The results are shown in Table.~\ref{different-interval}. We can observe that the proposed method performs better than the best baseline DSen2-CR overall, which is consistent with the results on the SEN12MS-CR dataset. It shows the feasibility of assessing the performances with temporally close cloud-free images. And we can observe that, when the interval between the reference cloud-free image used to calculate the value of the metrics and the cloudy image is larger, the methods performs worse in terms of the metrics. It indicates that the method is able to restore the surface information of the input cloudy image, and thus the cloud-free image with the larger interval to input cloudy image has less reference value.
    
\begin{table}[!h]
\small
\centering
\caption{Evaluating cloud removal performance using the cloud-free images with different intervals from cloudy images.}
\label{different-interval}
\resizebox{.49\textwidth}{!}{
    \begin{tabular}{cccccc}
    \hline
    Interval                & Method   & PSNR $\uparrow$& SSIM $\uparrow$& SAM $\downarrow$& MAE $\downarrow$ \\ \hline
    \multirow{2}{*}{13 days} & DSen2-CR &27.6299      &0.8618     &6.9426      &0.0293\\
                            & GLF-CR   &28.6470      &0.8707     &6.9005      &0.0260\\ \hline
    \multirow{2}{*}{25 days} & DSen2-CR &26.3796      &0.8403     &8.0728      &0.0334\\
                            & GLF-CR   &26.9173      &0.8444     &8.6355      &0.0317\\ \hline
    \multirow{2}{*}{72 days} & DSen2-CR &25.1544      &0.8247     &10.0843      &0.0382\\
                            & GLF-CR   &25.3110      &0.8324     &10.6852      &0.0378\\ \hline
    \end{tabular}
}
\end{table}

Strict ground truth correspondence may only be guaranteed by generating synthetic cloud coverage superimposed on cloud-free observations, as done in \cite{enomoto2017filmy} and \cite{gao2020cloud}. However, the experimental results in \cite{ebel2020cloud} has indicated that popular synthetic cloud simulation techniques suffer from severe limitations in approximation to the real data. The great performance on synthetic data may not necessarily translate to equal performance on real data. Hence we follow the approach of using real observations, despite acknowledgeable shortcomings at other ends.

\section{Conclusion}
In this work, we propose a novel global-local fusion based cloud removal (GLF-CR) algorithm for high quality cloud-free image reconstruction.  
It boosts cloud removal performance from two aspects, on the one hand, it guides the relationship among all local optical windows with the SAR feature to fully utilize the spatial consistency between the cloudy and the neighboring cloud-free regions, and on the other hand, it enhances the utilization of SAR data to compensate for missing information while alleviating the performance degradation caused by speckle noise. Extensive experiments demonstrate that the power of the information embedded in neighboring cloud-free regions and corresponding SAR data over different cloud cover levels. The proposed method can achieve state-of-the-art performance on all different cloud cover levels.

\section*{Acknowledgements}

F. Xu is supported by the China Scholarship Council (CSC). The work of W. Yang is supported by the National Natural Science Foundation of China (NSFC) under Grant 61771351. The work of X. Zhu is jointly supported by the European Research Council (ERC) under the European Union's Horizon 2020 research and innovation programme (grant agreement No. [ERC-2016-StG-714087], Acronym: \textit{So2Sat}), by the Helmholtz Association
through the Framework of Helmholtz AI (grant  number:  ZT-I-PF-5-01) - Local Unit ``Munich Unit @Aeronautics, Space and Transport (MASTr)'' and Helmholtz Excellent Professorship ``Data Science in Earth Observation - Big Data Fusion for Urban Research''(grant number: W2-W3-100), by the German Federal Ministry of Education and Research (BMBF) in the framework of the international future AI lab "AI4EO -- Artificial Intelligence for Earth Observation: Reasoning, Uncertainties, Ethics and Beyond" (grant number: 01DD20001) and by German Federal Ministry of Economics and Technology in the framework of the "national center of excellence ML4Earth" (grant number: 50EE2201C).











\bibliographystyle{cas-model2-names}

\bibliography{cas-refs}



\end{document}